\documentclass[lettersize,journal]{IEEEtran}
\usepackage{amsmath,amsfonts}
\usepackage{algorithmic}
\usepackage{algorithm}
\usepackage{array}
\usepackage[caption=false,font=normalsize,labelfont=sf,textfont=sf]{subfig}
\usepackage{textcomp}
\usepackage{stfloats}
\usepackage{url}
\usepackage{verbatim}
\usepackage{graphicx}
\usepackage{cite}
\usepackage{soul, color}

\usepackage[colorlinks,
            linkcolor=blue,
            anchorcolor=blue,
            citecolor=blue]{hyperref}
            
\usepackage{multirow}
\usepackage[table,xcdraw]{xcolor}
\usepackage[normalem]{ulem}
\useunder{\uline}{\ul}{}
\usepackage{amssymb}

\begin{document}

\title{MPT-PAR:Mix-Parameters Transformer for Panoramic Activity Recognition}

\author{Wenqing Gan, Yan Sun, Feiran Liu, Xiangfeng Luo
\thanks{This work was supported in part by the National Natural Science Foundation of China under Grant 62002215, and in part by the Shanghai Pujiang Program under Grant 20PJ1404400.}
\thanks{Wenqing Gan, Yan Sun, Feiran Liu and Xiangfeng Luo are with the School of Computer Engineering and Science, Shanghai University, Shanghai 200444, China (e-mail: ganwenqing@shu.edu.cn; yansun@shu.edu.cn; shirleyllfr@shu.edu.cn; luoxf@shu.edu.cn).}}

\markboth{Journal of \LaTeX\ Class Files,~Vol.~14, No.~8, August~2021}%
{Shell \MakeLowercase{\textit{et al.}}: MPT-PAR:Mix-Parameters Transformer for Panoramic Activity Recognition}

\IEEEpubid{0000--0000/00\$00.00~\copyright~2021 IEEE}

\maketitle

\begin{abstract}
The objective of the panoramic activity recognition task is to identify behaviors at various granularities within crowded and complex environments, encompassing individual actions, social group activities, and global activities. Existing methods generally use either parameter-independent modules to capture task-specific features or parameter-sharing modules to obtain common features across all tasks. However, there is often a strong interrelatedness and complementary effect between tasks of different granularities that previous methods have yet to notice. In this paper, we propose a model called MPT-PAR that considers both the unique characteristics of each task and the synergies between different tasks simultaneously, thereby maximizing the utilization of features across multi-granularity activity recognition. Furthermore, we emphasize the significance of temporal and spatial  information by introducing a spatio-temporal relation-enhanced module and a scene representation learning module, which integrate the the spatio-temporal context of action and global scene into the feature map of each granularity. Our method achieved an overall $F_1$ score of 47.5\% on the JRDB-PAR dataset, significantly outperforming all the state-of-the-art methods.
\end{abstract}

\begin{IEEEkeywords}
Human Activity Understanding, Group Activity, Panoramic Activity Recognition.
\end{IEEEkeywords}

\section{Introduction}
\IEEEPARstart{D}{ue} to the extensive application of human activity recognition tasks in various fields such as video surveillance \cite{7514916,sun2022human,yu2022argus++}, healthcare \cite{zouba2009assessing,haque2017towards,bates2021potential} and sports analysis \cite{li2023dual,askari2022interaction,10159436}, this task has become crucial in video understanding. Previously, human activity recognition primarily focused on single-granularity tasks, such as recognizing individual actions \cite{xing2023svformer,wu2023can,chen2023agpn} or group activities \cite{pei2023key,zhou2022composer,chappa2023spartan}. These tasks require a fixed number of people in the scene and assume that, in multi-person scenes, all individuals belong to a single group without any subdividable subgroups. These assumptions are often unrealistic in real-world scenarios, thereby limiting the utility of these methods. Moreover, focusing on single-granularity tasks makes it difficult to understand the entire scene, while tasks of different granularities can mutually reinforce each other. For instance, accurate individual actions recognition can help the model learn the commonalities of individual behaviors, thereby aiding in more accurate group activity recognition. To address these issues, the panoramic activity recognition task, as proposed by \cite{han2022panoramic}, aims to establish a unified framework capable of recognizing behaviors at different granularities simultaneously in crowded and complex scenes, including individual actions, social group activities, and global activities. This task does not impose restrictions on the number of people in the scene and allows for the subdividing groups. Recognizing behaviors of different granularities within a unified framework leverages the complementary relationships between tasks, as different granularities introduce diverse inductive biases that broaden the model’s perspective, enhancing its generalization ability to unknown data distributions.

To establish a unified framework for simultaneous multi-task learning, the most critical factor is the structure design—specifically, determining which parts are task-specific and which are shared across tasks. This significantly impacts performance, training efficiency, and generalization. Current methods utilize parameter-sharing backbones and individual feature relationship learning modules to learn the foundational features common to multi-granularity activity recognition tasks, and then use multiple task-specific parameter-independent classification heads for multitask recognition. For cross-granularity aggregation modules, JRDB-PAR \cite{han2022panoramic} uses parameter-independent modules to better capture specific features and nuances of each task, avoiding optimization conflicts of different task supervision signals; on the other hand, MUP \cite{cao2023mup} uses parameter-sharing modules to extract common relationships and motion features, promoting information sharing and enhancing overall performance. However, these methods do not effectively to simultaneously leverage the advantages of both aggregation strategies and overlook their complementary effects. Besides, since social groups and global activities are aggregated from individual actions, leveraging temporal information of individuals can lead to a better understanding of the scenarios. However, this requires multi-object tracking and group evolution detection, and they are very difficult to implement in challenging panoramic scenes. Therefore, previous methods ignored temporal information and relied on single-frame recognition, resulting in suboptimal performance. Furthermore, previous methods performed panoramic activity recognition using only cropped individual features, limiting the features to a local receptive field. As a result, they inevitably neglect the advantages of the global contextual information.

In this work, we propose a novel network called Mix-Parameters Transformer for Panoramic Activity Recognition (MPT-PAR) to address those challenges. \IEEEpubidadjcol A cross-granularity aggregation module is introduced to integrate features from individuals to social groups and from individuals to the global level. This module, comprising structurally identical parameter-independent and parameter-sharing encoders, models relationships between individuals through multi-head self-attention and aggregates individual behavior information via a learnable cls token. The fused features from both encoders are combined, to capture common relationships and motion features while retaining specific task features and nuances. To exploit temporal information, we follow the conventional approach for multi-object tracking and group evolution detection in group activity recognition, which is closely related to panoramic activity recognition. We also propose spatio-temporal relation-enhanced module, which model spatial relationships between different individuals in the same frame and temporal relationships of the same individual across frames. To provide comprehensive global context, we expand the receptive field by modeling scene information of the entire video frame, capturing behavior-related elements through spatial attention, flexibly integrating scene representations and activity features through cross-attention. Extensive experiments are conducted on the JRDB-PAR dataset to validate the effectiveness of the proposed network. Our method achieved an overall $F_1$ score of 47.5\%, significantly improving by 6.0\% compared to state-of-the-art methods. Additionally, it shows the greatest improvement in the sub-task of global activity recognition compared to previous methods, with an $F_1$ score of 61.1\%. We also compared our approach with SOTA methods that perform single-granularity tasks, the significantly improvement demonstrating the mutual reinforcement between tasks of different granularities.

The main contributions of this paper are:

\begin{itemize}
    \item We propose a novel network MPT-PAR, it uses both parameter-sharing and parameter-independent aggregation encoders for cross-granularity feature integration, leveraging the complementary advantages to improve panoramic activity recognition.
    \item We introduce spatio-temporal relation-enhanced module to utilize temporal information by enhancing spatial and temporal relationships for individual features.
    \item We explicitly model scene information to provide comprehensive global context, captures behavior-related elements through spatial attention, and integrate scene representations with activity features through cross-attention.
    \item Conducting extensive experiments on the JRDB-PAR dataset, we demonstrate the effectiveness of proposed method with an overall $F_1$ score of 47.5\%, showing a significant improvement of 6.0\% over state-of-the-art methods.
\end{itemize}

\section{RELATED WORK}

\subsection{Single Granularity Activity Recognition}
\subsubsection{Human Action Recognition}
Human action recognition (HAR) is a critical research area in computer vision that aims to automatically identify various human actions from video data. The advancement of deep learning has led to the development of numerous models based on Convolutional Neural Networks (CNN) \cite{simonyan2014two,tran2018closer,kalfaoglu2020late} and Recurrent Neural Networks (RNN) \cite{liu2016spatio,majd2020correlational,Baradel_2017_ICCV} for this purpose. A milestone in this field is the Two-Stream Convolutional Network \cite{simonyan2014two}, which utilizes static frame images and optical flow images to separately capture spatial and temporal features, resulting in significant performance gains. Additionally, RNN architectures like Long Short-Term Memory (LSTM) \cite{liu2016spatio,majd2020correlational} and Gated Recurrent Units (GRU) \cite{Baradel_2017_ICCV} have been extensively applied to video sequence data, further improving action recognition accuracy. In recent years, Transformer-based frameworks \cite{arnab2021vivit,fan2021multiscale,10262342} have been incorporated to enhance the model's ability to focus on key frames and regions, thereby boosting recognition performance. These techniques capture intricate spatio-temporal relationships and feature interactions, achieving state-of-the-art performance on several public datasets. Moreover, multimodal fusion approaches, such as networks combining depth information \cite{ji2020arbitrary} and skeletal data \cite{bruce2022mmnet}, have shown superior performance in complex scenarios.

\subsubsection{Social Group Detection and Activity Recognition}
The task of social group detection aims to divide people into subgroups based on their social activities or relationships. Building on this concept, social activity recognition aims to simultaneously identify the social activities of each subgroup. Early methods focused primarily on social group detection due to hardware limitations and small datasets. Solera et al. \cite{7214317} manually designed features representing body and social identity to understand relationships among group members and used trajectory clustering to detect social groups. Wang et al. \cite{wang2020panda} introduced a benchmark for social group detection with a “global to local magnification” framework that encodes global trajectories and local interactions, yielding promising results. Recently, the focus has shifted towards social activity recognition. Ehsanpour et al. \cite{ehsanpour2020joint} developed an end-to-end trainable network using integrated self-attention and graph attention modules to recognize social subgroups and their behaviors. However, this method oversimplifies social activities by treating them as the dominant individual actions within each group, which is not practical in many real-world scenarios. To address this, Kim et al. \cite{kim2023practicalgroupactivitydetection} created a large-scale high-definition dataset specifically for social behavior detection and proposed a new Transformer-based model capable of handling an unknown number of groups and potential group members without relying on clustering algorithms.

\subsubsection{Group Activity Recognition}
It is a subfield of human activity recognition, focuses on identifying activities involving multiple individuals. The early work in this field primarily relied on manually designed features to extract information from video frames. Hierarchical graphical models \cite{amer2014hirf} and dynamic Bayesian networks \cite{zhu2013context} were commonly used to interpret group activities in video environments. With the advent of deep learning, recent approaches have favored using deep neural networks for group activity recognition tasks. Yuan et al. \cite{DBLP:journals/corr/abs-2108-11743} proposed a GCN-based method through a Dynamic Relation module and Dynamic Walk module for spatio-temporal individual inference. Xie et al. \cite{Causality_Graph} introduced an actor-centric causality graph that focuses on analyzing the impact of two actors on asynchronous causality detection, complementing the learned asynchronous relationships with the synchronous ones derived from a Transformer model. GroupFormer \cite{li2021groupformer} employs a clustering attention mechanism, leveraging a clustering space-time Transformer to enhance individual and group representations by integrating spatial and temporal dependencies. Addressing the reliance on bounding box annotations for group activity recognition, a detector-free Transformer-based model called DFWSGAR \cite{kim2022detector} was proposed, which locates and encodes parts of the group activity context through attention mechanisms, aggregating them into a single group representation without requiring bounding box labels or object detectors.

Overall, these methods focus solely on single-granularity tasks and they impose some unrealistic limitations. For instance, human action recognition often requires videos to contain only one or a few individuals, with the actors occupying the main part of the frame in each scene. In group activity recognition, the number of people in the video is usually fixed, and all individuals are assumed to belong to a single group without subdivisions. Those unrealistic assumptions significantly limit the application of these methods. Additionally, single-granularity task models did not realize the mutual reinforcement between tasks at different levels. In contrast, the objective of this work is to simultaneously achieve individual action recognition, social group activity recognition, and global activity recognition in crowded scenes, making it more practical for many real-world applications.

\subsection{Panoramic Activity Recognition}
Panoramic activity recognition aims to identify individual actions, social group activities, and global activities concurrently. As a newly proposed and challenging problem, this field remains underexplored with limited current work. Han et al. \cite{han2022panoramic} proposed a novel hierarchical graph neural network to progressively represent and model multi-granularity activities and mutual social relationships within crowds. By modeling each individual as a graph node, a graph network is constructed. Multiple parameter-independent GCN-based cross-granularity aggregation modules aggregate individual feature nodes into social group nodes bottom-up. Similarly, social group nodes are further aggregated into global nodes. This parameter-independent aggregation approach better captures the specific features and nuances of each task, avoiding optimization conflicts in supervising signals on the same parameters and improving individual task performance. In contrast, MUP \cite{cao2023mup} uses a unified cross-granularity aggregation module to encode behaviors at different granularity levels with the same parameters in an end-to-end manner. This approach learns potentially generic motion patterns across multi-granularity human behaviors, facilitating information sharing among tasks and enhancing the performance of panoramic activity recognition sub-tasks.

However, these methods only use either parameter-independent or parameter-sharing modules to integrate cross-granularity activity features, overlooking the complementary strengths of these two methods. Besides, these methods perform activity recognition at each granularity based solely on individual features extracted from a single key frame, without considering temporal information and the overall scene context.

\section{PROPOSED APPROACH}
\subsection{Overview}
\begin{figure*}[h]
\centering
\includegraphics[width=7in]{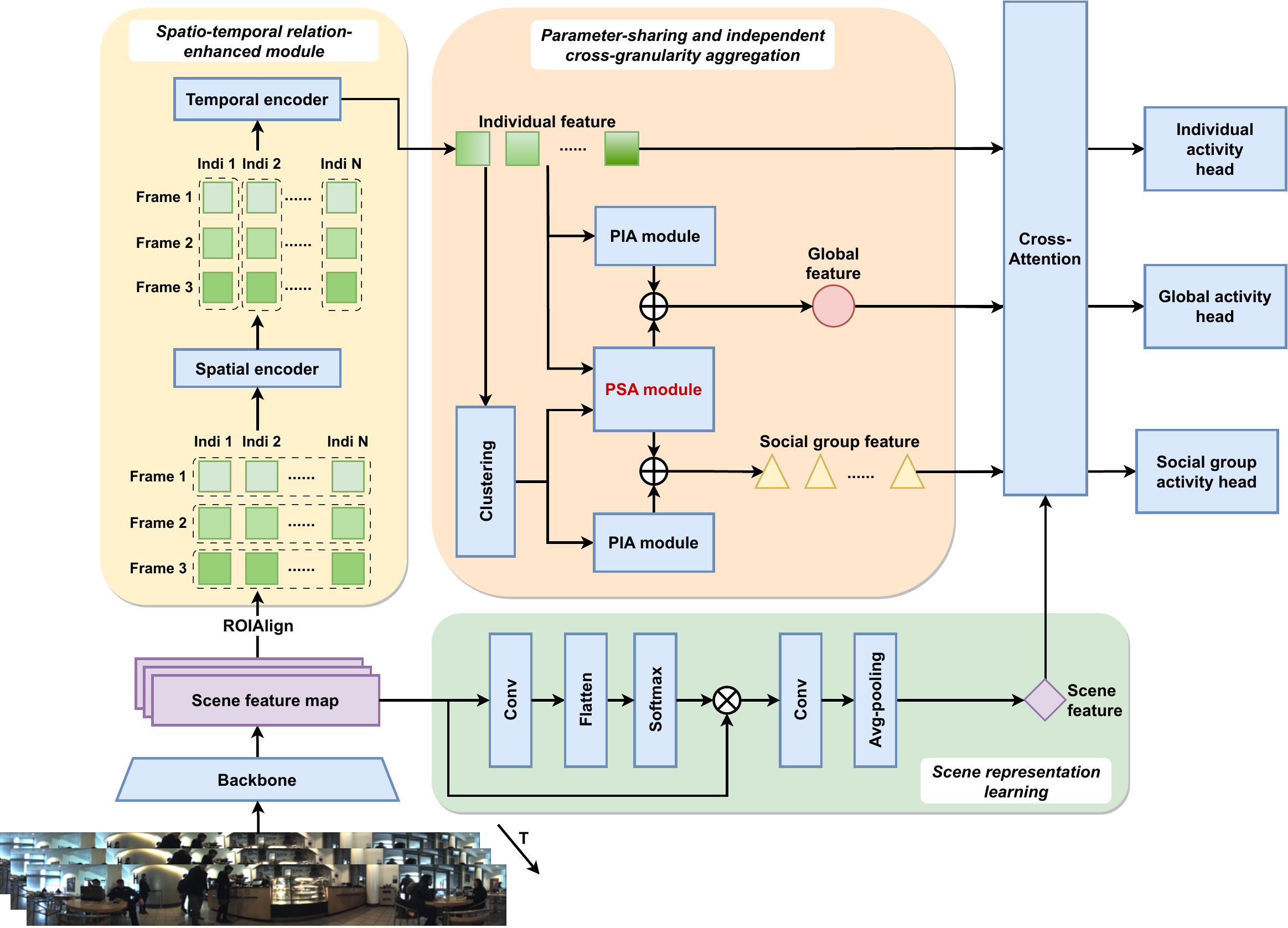}
\caption{Overview of our proposed model.} \label{fig1}
\end{figure*}
The architecture of our network is illustrated in Fig. \ref{fig1}. The backbone is a ResNet-18 network \cite{he2016deep} pre-trained on ImageNet. For the input video frame sequence $X_{img} \in R^{T\times3\times H\times W}$ ($T$ indicates the number of frames), the backbone extracts feature maps resulting in scene features  $X_{scene} \in R^{T\times C\times H'\times W'}$ ($C$ means the number of channels). Each individual's features $X \in R^{T \times N\times D}$ ($N$, $D$ represent the number of individuals and the dimension of individual's features respectively) are then cropped using RoIAlign \cite{he2017mask}. The scene features $X_{scene}$ are passed through a scene representation learning module to obtain the scene representation. The cropped individual features $X$ are enhanced by spatio-temporal relation-enhanced module (STRE), which enrich individual features with spatio-temporal relational context. Subsequently, parameter-sharing and independent cross-granularity aggregation module (PSICGA) is employed to integrate individual features into social group and global representations. Before classifying the individual, social group, and global features, they are fused with scene representations through cross-attention to fully exploit the scene context.

\subsection{Spatio-Temporal Relation-Enhanced Module (STRE)}

 Previous methods ignored temporal information and relied solely on single frames for panoramic human activity recognition. However, actions are typically executed over a sequence of continuous frames. Ignoring temporal information can result in the loss of action details, leading to inaccurate understanding of actions by the model, which in turn affects the recognition of social group activity and global activity. Moreover, in real-world scenarios, interpersonal relationships are dynamically changing, making it unreliable to infer groupings of people based on a single image.
 
 To address this issue, we introduce temporal information and use spatio-temporal relation-enhanced module to obtain features enriched with temporal and spatial context. Our STRE module consist of a transformer-based \cite{dosovitskiy2020image} temporal encoder and a spatial encoder, which learn temporal and spatial individual contextual features, respectively. Inspired by TimeSformer \cite{bertasius2021space}, we connect the temporal and spatial encoders serially to reduce computational complexity. Experiments demonstrated that the serial structure is simpler and more effective (details in Section \ref{section:STRE}).

\subsubsection{Spatial Encoder}
In crowded multi-person scenes, individual actions are not entirely independent but are significantly influenced by interactions with others and the behaviors of those present in the scene. Therefore, considering spatial context is crucial for panoramic human activity recognition. We employ a transformer-based spatial encoder to learn spatial contextual information of individuals. The spatial encoder captures interaction relationships between the target individual and surrounding individuals by assigning relative importance weights.

Given the input individual representation $X \in  R^{T \times N\times D}$, we treat the temporal dimension as the batch dimension and use different learnable projection matrix to map $X$ to $Q$, $K$ and $V$. The spatial relationship attention weights of neighboring individuals are then computed using $softmax$, and the weights are multiplied by $V$ to obtain the spatial context-enhanced feature $X_S\in  R^{T \times N\times D}$. For the $t$-th frame, this process can be expressed as:
\begin{equation}\label{eqn1} 
Q^{t}=X^tW_{tq},K^{t}=X^tW_{tk},V^{t}=X^tW_{tv}
\end{equation}

\begin{equation}\label{eqn2} 
    X_s^t = FFN(softmax(\frac{Q^t{K^t}^T}{\sqrt{D}})V^t+V^t)
\end{equation}where $W_{tq}$, $W_{tk}$ and $W_{tv}$ are learnable projection matrices, and $FFN$ is a feedforward network.

\subsubsection{Temporal Encoder}

Considering the temporal correlation of panoramic activities, the temporal encoder captures temporal contextual information by computing the adaptive importance of individuals across consecutive frames, enriching the temporal features of individuals. The structure of the temporal encoder is identical to that of the  spatial encoder. The difference lies in treating the spatial dimension as the batch dimension for the temporal encoder. We transpose the temporal and spatial dimensions of $X_s$, reshaping it to $X'_s \in  R^{N\times T \times D}$, and feed it into the temporal encoder, resulting in the time-context-enhanced individual representation $X_t\in  R^{N\times T \times D}$. As temporal information is only considered at the individual level, we average the time dimension of the enhanced individual features to obtain $X_{st}\in  R^{N\times D}$, serving as the foundation for constructing social group and global features.

\subsection{Parameter-Sharing and Independent Cross-Granularity Aggregation Module (PSICGA)}

As a multi-granularity and multi-task learning task, panoramic activity recognition requires simultaneously addressing individual action recognition, social group activity recognition, and global activity recognition. Using parameter-independent aggregation modules for bottom-up cross-granularity feature integration helps capture task-specific features relevant to each task and avoids optimization conflicts in shared parameters, thereby enhancing multi-task learning efficiency. However, as indicated by MUP \cite{cao2023mup}, different parameter modules model activities at various granularities separately, making it impossible for the model to learn potential general motion patterns across different granularities of human behavior. Therefore, we design a module that integrates parameter-sharing and parameter-independent mechanisms to learn task-specific features independently and alleviate optimization conflicts between tasks, while capturing common behavioral patterns during cross-granularity aggregation through shared parameters.

\begin{figure}[!t]
\centering
\includegraphics[width=3in]{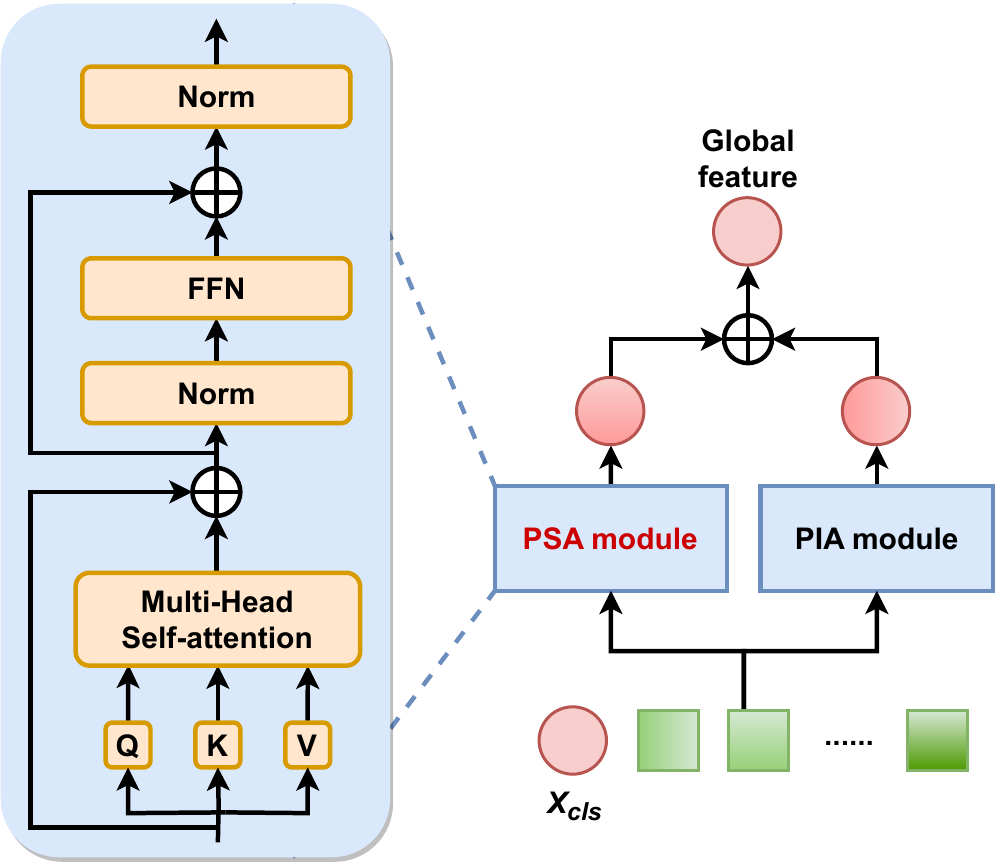}
\caption{Structure of PSICGA module.} \label{fig2}
\end{figure}

As shown in Fig. \ref{fig2}, the proposed PSICGA module is constructed from two encoders: a parameter-sharing aggregation encoder (PSA) and a parameter-independent aggregation encoder (PIA). Although these two encoders share the same structure, which incorporates layer normalization, multi-headed self-attention and FFN, the parameters in them are different. The PSICGA module employs a parallel structure, where the input features are fed into both the PSA and PIA encoder. The network then combines the cls tokens from the outputs of the two modules to obtain the fused features.

For individual to global aggregation, we first add a learnable cls token $X_{cls}\in  R^{1\times D}$ to the individual feature sequence $X_{st}\in  R^{N\times D}$. It summarizes the individual behavior context and the interactions learned by the multi-headed self-attention, representing the aggregated global activity features. To better model the relationships within the aggregation module, we add learnable spatial position embedding to provide positional information. The individual feature sequence $X_{st}'\in  R^{(N+1)\times D}$ with the added cls token and spatial position encoding  is fed into the PIA encoder $PIA_{global}()$ and PSA encoder $PSA()$. The cls tokens representing global activity features from the two modules are combined to obtain the cross-granularity aggregated global activity feature $X_{global}$. The process is defined as:

\begin{equation}\label{eqn3}
    X_{st}'=[X_{cls},x_{st}^1,x_{st}^2...x_{st}^n]+P
\end{equation}

\begin{equation}\label{eqn4}
    X_{cls}^{global} = PIA_{global}(X_{st}'),X_{cls}^{share} = PSA(X_{st}')
\end{equation}

\begin{equation}\label{eqn5}
    X_{global}=X_{cls}^{global}+\lambda X_{cls}^{share}
\end{equation}
where $P$ represents spatial position embedding, and $\lambda$ is a weight parameter controlling the addition. 

For individual to social group integration, we first use clustering to obtain social group detection results, then perform cross-granularity aggregation for each social group. The PSICGA module used in this aggregation is similar to the one used in individual to social group integration, with the only difference being that the PIA encoder with different parameters $PIA_{social}()$ is used to obtain the corresponding $X_{cls}^{social}$, while $PSA()$ encoder is reused for both integrations.

\subsection{Scene Representation Learning}

The global scene contextual information is crucial for panoramic activity recognition. Our scene representation learning module consists of scene representation generation and scene representation fusion. First, visual scene tokens are generated from the scene features extracted by the backbone, and they are aggregated to form the scene representation. Then, cross-attention is employed to fuse the scene representation with activity features at different granularities.
\subsubsection{Scene Representation Generation}
Inspired by Visual Transformers \cite{wu2020visual}, the global scene information in video frames can be summarized using a compact set of visual tokens. We use a set of visual scene tokens to summarize the global scene information of the current video frame and aggregate them into a scene representation. For the scene feature map $X_{scene} \in  R^{T\times D\times H'\times W'}$ extracted by the backbone, we first add learnable positional encodings to retain positional information. Next, pointwise convolution is applied to assign each pixel to $K$ visual scene tokens $X_{scene}' \in  R^{T\times K\times (H'·W')}$. A spatial attention matrix $A\in R^{T \times  K \times  ( H'· W')}$ is generated using $softmax$, and the spatial attention weights are used to compute the weighted sum of each pixel. 2D convolution is applied to aggregate the scene features along the spatial dimension $( H'· W')$ to channel $D$, generating $K$ visual scene tokens $Z\in R^{T \times  K \times  D}$. This process can be described as:
\begin{equation}\label{eqn6}
    Z=conv(softmax(conv(X_{scene}))X_{scene})
\end{equation}
Through the spatial attention mechanism, each visual scene token adaptively focuses on notable parts of the entire scene. Finally, we perform AvgPooling on the first two dimensions of the visual scene tokens, merging the $K$ visual scene tokens into the scene representation $X_{S} \in  R^{1\times D}$.

\subsubsection{Scene Representation Fusion}
To fully integrate the global context information contained in the scene representation with features at various granularities, we use a multi-head cross-attention and a feedforward network for fusion. For individual-level features, the inputs to the multi-head cross-attention are the enhanced individual representation $X_t\in  R^{N\times 1 \times D}$ and the scene representation $X_{S} \in  R^{1\times D}$. The individual representation is used as the query, and the scene representation  as the key in the multi-head cross-attention to fuse individual and scene representations, yielding the global scene context enhanced individual representation. This operation is similarly performed at the social group level. In the global activities recognition task, we directly concatenate the global-level features with the scene representation and use a learnable MLP to fuse them.

\subsection{Multi-Task Loss Function} 

Following JRDB-PAR \cite{han2022panoramic}, we sum the losses of multiple tasks to serve as supervision for panoramic activity recognition. The total loss is defined as:
\begin{align}\label{eqn7}
\mathcal{L} &= \mathcal{L}_i + \mathcal{L}_s + \mathcal{L}_g + \mathcal{L}_d \notag \\ &= \sum_u L(\mathbf{a}^I_u, \tilde{\mathbf{a}}^I_u) + \sum_k L(\mathbf{a}^S_k, \tilde{\mathbf{a}}^S_k) + L(\mathbf{a}^G, \tilde{\mathbf{a}}^G) + L(\mathbf{R}, \tilde{\mathbf{R}})
\end{align}
where $\mathcal{L}_i$, $\mathcal{L}_s$, $\mathcal{L}_g$, and $\mathcal{L}_d$ represent the losses for individual, social group, global activity recognition, and social group detection, respectively. $u$ and $k$ denote each individual's features and social group features in the scene. $L(\cdot)$ represents binary cross-entropy loss, and $\tilde{\cdot}$ denotes the ground truth labels. $R$ is the interpersonal relationship matrix with values of 0 or 1, where 1 indicates that the corresponding two individuals belong to the same group.

\section{Experiment}

\begin{table*}[t]
\centering
\caption{Comparative results of the panoramic activity recognition (\%). The best results are marked in \textbf{bold} and the second best ones are \underline{underlined}. “AR”, “GAR”, and “PAR” represent Action Recognition, Group Activity Recognition and Panoramic Activity Recognition separately. DFWSGAR \cite{kim2022detector} and DECOMPL \cite{demirel2023decompl} do not employ individual action recognition as an auxiliary task, so their results for individual action recognition cannot be compared.}\label{tab1}
\begin{tabular}{p{0.3cm}c|ccc|ccc|ccc|c}
\hline
\multicolumn{2}{c}{\multirow{2}{*}{Method}}                     & \multicolumn{3}{c}{Individual Act.} & \multicolumn{3}{c}{Social Group Act.} & \multicolumn{3}{c}{Global Act.} & Overall \\
\multicolumn{2}{c}{}                                            & $P_i$          & $R_i$          & $F_i$          & $P_p$          & $R_p$           & $F_p$           & $P_g$         & $R_g$         & $F_g$        & $F_a$      \\ \hline
\multirow{2}{0.3cm}{A R}             & UniFormer\cite{li2022uniformer}& 51.0       & 40.5        & 43.7       & -           & -            & -            & -          & -          & -         & -       \\
                                                & vid-TLDR\cite{choi2024vid}      & 53.0       & 43.4        & 45.7       & -           & -            & -            & -          & -          & -         & -       \\ \hline
\multirow{5}{0.3cm}{G A R}     & ARG\cite{wu2019learning}           & 39.9        & 30.7        & 33.2        & -           & -            & -            & 63.6       & 44.3       & 50.7      & -    \\
                                                & SA-GAT\cite{ehsanpour2020joint}        & 44.8        & 40.4        & 40.3        & -           & -            & -            & 36.7       & 29.9       & 31.4      & -    \\
                                                & JRDB-Base\cite{ehsanpour2022jrdb}     & 19.1        & 34.4        & 23.6        & -        & -         & -         & 44.6       & 46.8       & 45.1      & -    \\ 
                                                &  DFWSGAR\cite{kim2022detector} & -           & -           & -           & -           & -            & -            & 57.2       & \textbf{66.4}      & 60.1     & -       \\
                                                & DECOMPL\cite{demirel2023decompl}       & -           & -           & -           & -           & -            & -            & 64.6      & 49.5      & 54.5     & -       \\ \hline
                                                
\multirow{3}{0.3cm}{P A R} 
                                                & JRDB-PAR\cite{han2022panoramic}      & 51.0          & 40.5        & 43.4        & 24.7        & 26.0           & 24.8         & 52.8       & 31.8       & 38.8      & 35.6    \\
                                                & MUP\cite{cao2023mup}           & 55.4        & 44.8        & 47.7        & 25.4        & 26.6         & 25.1         & 58.0         & 49.0         & 51.8      & 41.5    \\
                                                & Ours          & \textbf{59.2} & \textbf{58.6} & \textbf{56.0} & \textbf{25.5} & \textbf{27.3} & \textbf{25.4} & \textbf{69.1} & \underline{57.6}   & \textbf{61.1} & \textbf{47.5}   \\
\hline
\end{tabular}
\end{table*}

\subsection{Datasets and Metrics}
 The JRDB-PAR \cite{han2022panoramic} dataset is a novel benchmark specifically created for the panoramic human activity recognition task. This dataset extends the JRDB-Act \cite{ehsanpour2022jrdb} dataset by manually annotating social group activities and global activities, and adding annotations for social group detection. The JRDB-PAR dataset comprises 360-degree RGB videos captured by a mobile robot in various densely populated areas such as campuses, cafeterias, and classrooms, providing 628,000 annotated personal bounding boxes with IDs. The dataset employs multi-class labels for activity annotation, meaning each individual, group, or frame can have one or multiple activity labels. It includes 27 individual action classes (e.g., walking, talking), 11 social group activities (e.g., chatting, working together), and 7 global activities (e.g., commuting, conversing). The dataset consists of 27 video segments, totaling 27,920 frames, with 20 segments used for training and 7 for testing. Key frames are selected every 15 frames for annotation and evaluation.
 
 For individual actions recognition evaluation, we utilize common metrics in multi-label classification tasks: precision, recall, and $F_1$ score, to measure the accuracy of activity classification for each instance in the test dataset. Precision, recall, and $F_1$ score are denoted as $P_i$, $R_i$, and $F_i$ respectively. Social group activity recognition involves both group detection and activity category recognition. Group detection uses the classical Half metric \cite{wang2020panda} (IoU $>$ 0.5), where a detected group is considered correct if the intersection-over-union between predicted and actual group members exceeds 0.5. If the activity category of a detected group is correctly predicted, it is regarded as a correct social group activity prediction. Precision, recall, and $F_1$ score for social group activity recognition are denoted as $P_p$, $R_p$, and $F_p$ respectively. For global activity recognition, we also evaluate using precision, recall, and $F_1$ score, denoted as $P_g$, $R_g$, and $F_g$. The overall evaluation metric for panoramic activity detection is the average of these three $F_1$ scores, denoted as the comprehensive $F_1$ score $F_a=(F_i+F_p+F_g)/3$. This approach allows for a comprehensive assessment of individual actions detection, group activity recognition, and global activity recognition.

\subsection{Implementation Details}

 \subsubsection{Model Details}
    We use a ResNet-18 \cite{he2016deep} pretrained on the ImageNet dataset as the network backbone. In STRE, both the spatial and temporal encoders share the same structure, consisting of two layers with eight heads each. In PSICGA, the parameter-independent aggregation encoder has four layers with 12 heads, while the parameter-sharing aggregation encoder has two layers with 12 heads. For individual-to-social group aggregation, the weight parameter in Equation \eqref{eqn5} is $\lambda = 0.8$. For individual-to-global aggregation, $\lambda = 1$. The number of visual tokens for scene representation generation is set to 16.

 \subsubsection{Training}    
We implemented our network using the PyTorch framework and trained it on two GTX 3090 GPUs. Consistent with JRDB-PAR \cite{han2022panoramic}, we use ground truth social group detection labels during training and spectral clustering \cite{zelnik2004self} for social group detection during evaluation. We set the batch size to 2 and sample three frames. We use the Adam optimizer with an initial learning rate of $7 \times 10^{-6}$ and a weight decay of $10^{-2}$, training for 30 epochs.

\subsection{Comparison with State-of-the-Art Methods}

To demonstrate the superiority of our method, we compared it with state-of-the-art methods. Given the scarcity of existing panoramic activity recognition methods, we adapted the leading group activity recognition methods ARG \cite{wu2019learning}, SA-GAT \cite{ehsanpour2020joint}, JRDB-Base \cite{ehsanpour2022jrdb}, DFWSGAR \cite{kim2022detector}, and DECOMPL \cite{demirel2023decompl} for comparison. Since DFWSGAR \cite{kim2022detector} and DECOMPL \cite{demirel2023decompl} do not employ individual action recognition as an auxiliary task, their results for individual action recognition and social group activity recognition cannot be compared. Additionally, we applied the state-of-the-art individual activity recognition methods such as UniFormer \cite{li2022uniformer} and Vid-TLDR \cite{choi2024vid} to the JRDB-PAR dataset for comparison.

As shown in Table \ref{tab1}, our method outperforms the state-of-the-art panoramic activity recognition methods in all three sub-tasks and overall scores. Furthermore, compared to methods that focus on single-granularity tasks, our method exhibits superior performance across all levels of tasks, highlighting the mutual reinforcement among multiple tasks in panoramic activity recognition.

\begin{figure*}[t]
 \centering
\includegraphics[width=6in]{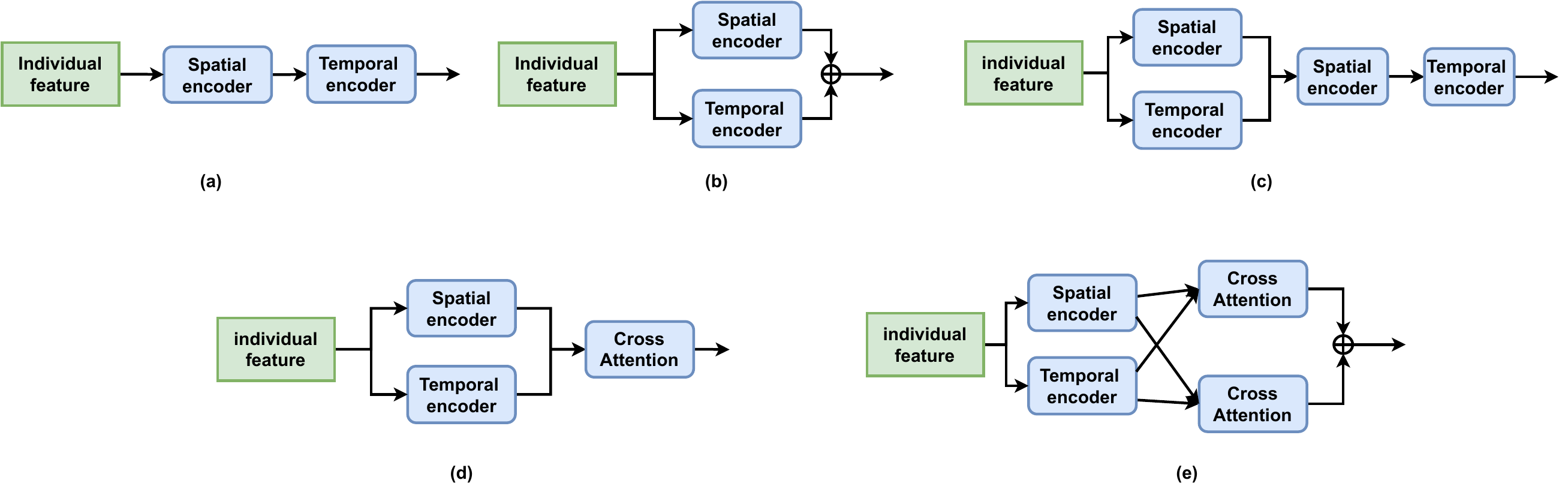}
\caption{The STRE structures studied in this work. (a) The serial structure. (b) The parallel structure. (c) The parallel-then-serial structure. (d) The structure with one cross-attention module. (e) The structure with two cross-attention modules.} \label{fig3}
\end{figure*}

\begin{table}
\centering
\caption{Ablation studies on STRE, PSICGA and SRL.}\label{tab2}
\begin{tabular}{ccc|c}
\hline
STRE & PSICGA & SRL & Overall $F_a$ \\ \hline
 & & &39.0\\ 
$\checkmark$    &&& 44.3     \\
&$\checkmark$      && 43.6      \\
&&$\checkmark$   & 41.7      \\
$\checkmark$     &$\checkmark$      && 46.9      \\
$\checkmark$     &&$\checkmark$   & 45.9     \\
&$\checkmark$      &$\checkmark$   & 44.4     \\
$\mathbf{\checkmark}$     &$\mathbf{\checkmark}$      &$\mathbf{\checkmark}$   & \textbf{47.5}      \\ \hline
\end{tabular}
\end{table}

\subsection{Ablation Studies}
\subsubsection{Effectiveness of Individual Modules}
We conducted experiments with different combinations of modules to verify their effectiveness. The experiment uses temporal average pooling on the collected individual features instead of STRE to evaluate the effectiveness of the module. Similarly for PSICGA, we replaced it with max pooling for cross-granularity aggregation. Additionally, we removed the scene representation learning (SRL) module for comparison. As shown in Table \ref{tab2}, when all the modules were removed, our baseline method still achieved a score of 39.0\%. This result surpasses some methods that only use single-frame information, indicating that even simple temporal average pooling can significantly enhance model performance. This highlights the importance of temporal information in panoramic activity recognition tasks. When the STRE was added, the model's performance further improved significantly, reaching a score of 44.3\%. Adding the PSICGA module and the SRL module individually also resulted in substantial improvements, demonstrating the effectiveness of our proposed modules. Subsequently, we combined the modules in pairs, and the synergy between the modules further improves the performance compared to using each module individually.

\subsubsection{STRE Module Structure}
\label{section:STRE}
 As shown in Fig. \ref{fig3}, we evaluated different structures of STRE, including serial structure (a), parallel structure (b), parallel-then-serial structure (c), and using one (d) and two (e) cross-attention mechanisms to fuse the temporal encoder and spatial encoder. Table \ref{tab3} shows that compared to the serial structure, the parallel structure's performance decreased significantly. The decline in performance may be due to the interference between the individual features caused by simultaneous temporal and spatial encoding, reducing overall performance.  Additionally, incorporating additional cross-attention to fuse parallel computation results was less effective than the serial structure. Our experimental results show that the serial structure used in STRE, despite its simplicity, better captures the spatio-temporal relationships of individual features, demonstrates its outstanding effectiveness.

\begin{table}
\centering
\caption{Ablation studies on STRE structure in figure \ref{fig3}.} \label{tab3}
\begin{tabular}{c|c}
\hline
Structure                           & Overall $F_a$ \\ \hline
Parallel (b)                           & 44.1     \\
Parallel-then-serial (c)               & 46.7      \\
One cross-attention module (d)   & 39.2       \\
Two cross-attention modules (e) &45.7\\
\textbf{Serial (a)} & \textbf{47.5} \\ \hline
\end{tabular}
\end{table}

\begin{table}
\centering
\caption{Ablation studies on PSICGA module structure.}\label{tab4}
\begin{tabular}{cc|c}
\hline

Parameter-independent & Parameter-sharing & Overall $F_a$    \\ \hline
$\checkmark$          & & 42.5          \\
& \textbf{$\checkmark$}      & 43.3          \\
$\checkmark$          & $\checkmark$      & \textbf{47.5} \\ \hline

\end{tabular}
\end{table}

\subsubsection{Effect of PSICGA Module}
To verify the effectiveness of this module, the network either uses parameter-sharing aggregation encoder or parameter-independent aggregation encoder individually for cross-granularity aggregation separately. As shown in Table \ref{tab4}, using the parameter-sharing aggregation encoder alone outperformed using the parameter-independent aggregation encoder alone. This might be because parameter sharing can better capture latent common behaviors across different levels, enhancing the model's generalization capability. Furthermore, when both parameter-sharing and parameter-independent aggregation encoders were used for cross-granularity aggregation, the network's performance significantly improved. This indicates that our proposed cross-granularity aggregation module can fully utilize the complementary effects of parameter sharing and independence. It can effectively learn task-specific features, alleviate parameter optimization conflicts, and capture latent common behaviors across different levels.
\begin{table}[]
\centering
\caption{Ablation studies on the number of visual scene tokens.}\label{tab5}
\begin{tabular}{c|c}
\hline
Numbers of token & Overall $F_a$ \\ \hline
4                & 40.6\%     \\
8                & 47.1\%     \\
\textbf{16}         & \textbf{47.5\%}     \\
24               & 46.3\%     \\
32               & 45.3\%     \\ \hline
\end{tabular}
\end{table}

\begin{figure}[t]
\centering
\includegraphics[width=0.3\textwidth]{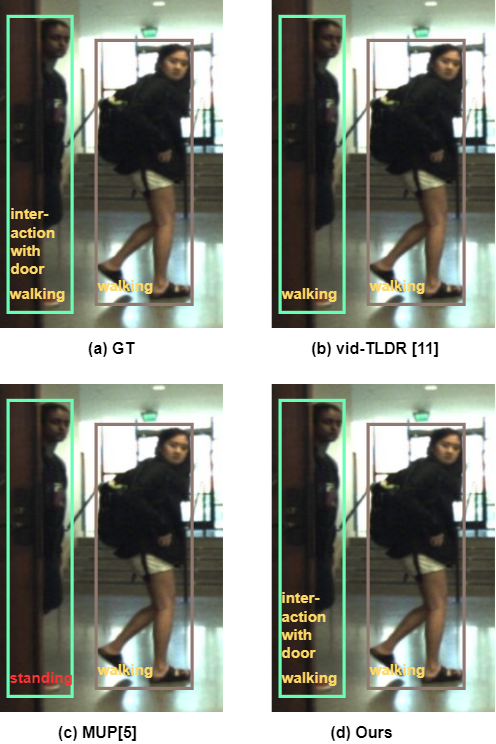}
\caption{The visualization of individual action granularity.} \label{fig4}
\end{figure}

\begin{figure}[t]
\centering
\includegraphics[width=0.4\textwidth]{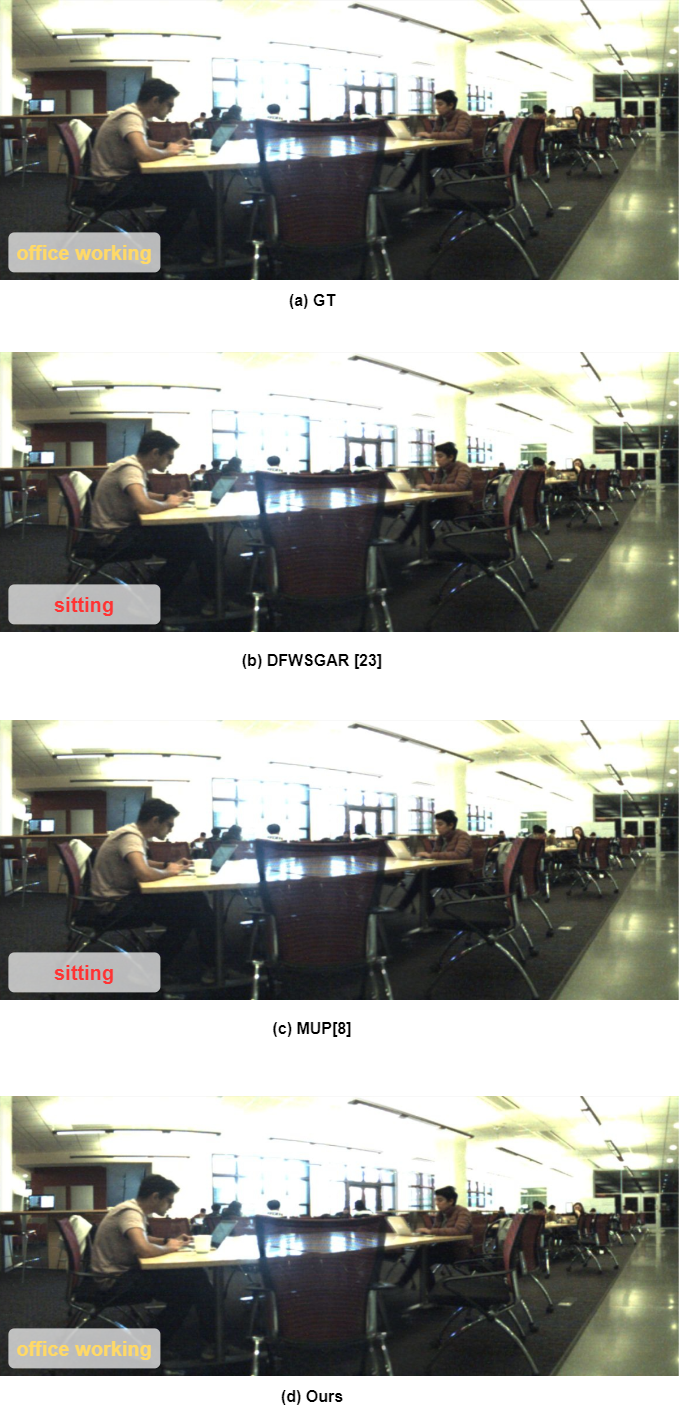}
\caption{The visualization of global activity granularity.} \label{fig5}
\end{figure}

\subsubsection{Investigation of the number of visual scene tokens}
We evaluated the impact of the number of visual scene tokens on experimental results in scene representation learning. The global scene information in video frames can be summarized using a small and compact set of visual scene tokens, with each token adaptively focusing on the important parts of the scene. Table \ref{tab5} shows that the model’s performance is gradually improved with the increasing  number of tokens from 4 to 16. More tokens help the model capture global scene information more effectively, thereby enhancing recognition accuracy. However, when the number of tokens exceeds 16, the model's performance starts to decline. This decline is likely due to information redundancy, which makes it harder for the model to extract useful information, thus negatively affecting performance.

\subsection{Qualitative Analysis}

\subsubsection{Visualization of Results}

\begin{figure*}
\centering
\includegraphics[width=0.7\textwidth]{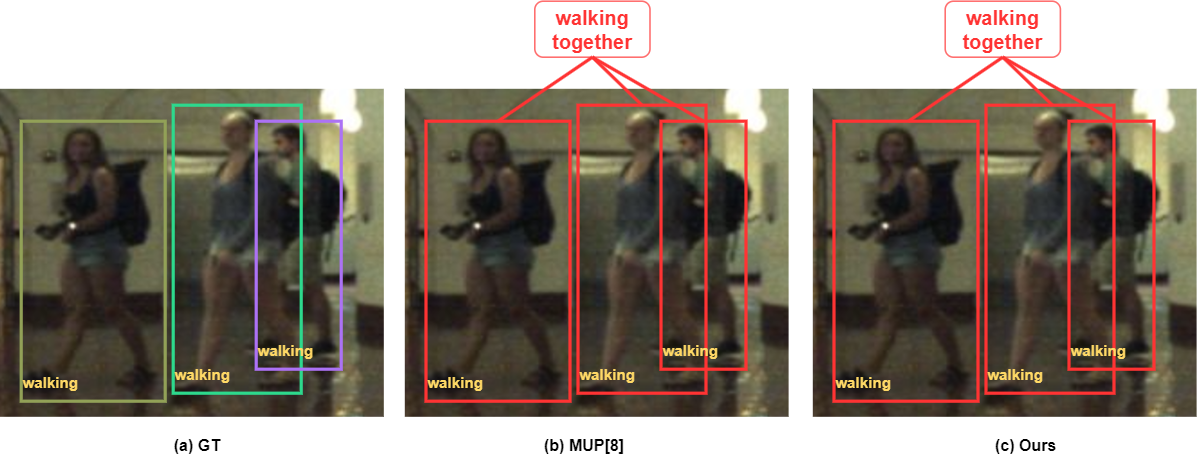}
\caption{The visualization of social group activity  granularity.} \label{fig6}
\end{figure*}

We visualized the prediction results of the MPT-PAR model and MUP \cite{cao2023mup} at different activity granularities. Besides, we compared our method with the vid-TLDR \cite{choi2024vid} and DECOMPL \cite{demirel2023decompl} at individual and global activity granularities. Fig. \ref{fig4} shows the results for individual action predictions. Since vid-TLDR \cite{choi2024vid} relies only on cropped individual features, it failed to correctly identify the “interaction with door” label. Additionally, MUP \cite{cao2023mup}, which overlooks temporal information, might struggle with occlusions, leading to incorrect identification of the label as “standing”. The comparison indicates that our model, utilizing temporal and scene information, accurately recognizes individual action even in crowded scenes with occlusions. The incorporation of global scene context provides additional scene information, enhancing recognition accuracy. As illustrated in Fig. \ref{fig5}, our model perceives scene semantics, such as computers, desks, and chairs, enabling more precise global behavior recognition.
However, the proposed model exhibits certain limitations. The direct application of spectral clustering \cite{zelnik2004self} for social group detection diminishes its accuracy. In Fig. \ref{fig6}, three individual pedestrians, who are in close proximity and performing the same “walking” action, are considered to belong to the same social group, therefore mistakenly thinking them “walking together”.
\begin{figure*}
\centering
\includegraphics[width=0.9\textwidth]{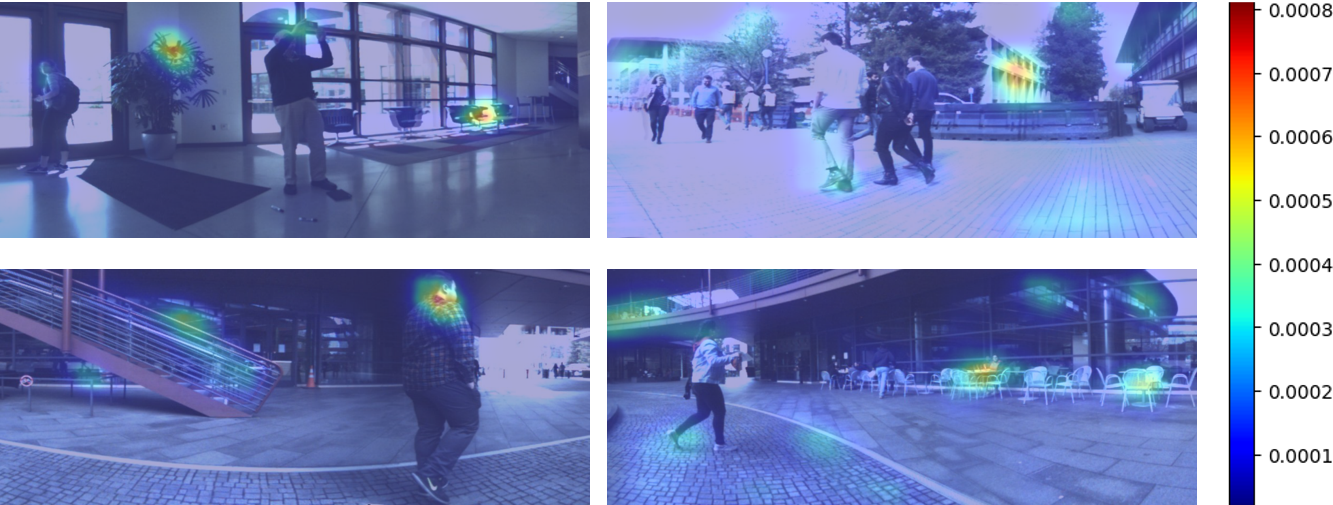}
\caption{The visualization of scene attention maps in the scene representation generation module.} \label{fig7}
\end{figure*}
\subsubsection{Visualization of Scene Attention Maps}
We visualized the spatial attention matrices computed in the scene representation learning module to better observe the distribution of attention within scenes. Fig. \ref{fig7} demonstrates that attention predominantly focuses on individuals within the scene, aligning with our expectations as individuals are the primary subjects of panoramic activity recognition. Additionally, significant attention is observed on other scene elements such as stairs, trees, railings, and chairs. These scene elements influence human behavior within the environment. For example, the presence of stairs correlates with actions of ascending or descending. The visualization results indicate that our scene representation module effectively captures contextual information relevant to behaviors, providing additional cues to enhance activity recognition in complex environments.

\section{Conclusion}
This paper introduces a novel network architecture, named MPT-PAR, for panoramic activity recognition. The network leverages complementary advantages of parameter sharing and independent cross-granularity aggregation modules, enhancing the effectiveness of activity recognition at all levels. Furthermore, temporal relationships enhanced by transformers and explicit modeling of scene information further help the model better understand the video content. Experiments on the JRDB-PAR dataset validate the efficacy of the proposed method, significantly improving performance across all granularity tasks. In future work, we will improve the reliability of social group aggregation and further enhance the practicality of the method.


\end{document}